\documentclass[11pt,a4paper]{article}
\usepackage[hyperref]{emnlp2020}
\usepackage{times}
\usepackage{latexsym}

\usepackage{times}
\usepackage{latexsym}
\usepackage{helvet}
\usepackage{amsmath}
\usepackage{courier}
\usepackage{subcaption}
\usepackage{graphicx}
\usepackage{url}
\usepackage{mathtools}
\usepackage{booktabs}
\usepackage{multirow}
\usepackage{amssymb}
\usepackage{graphicx}
\usepackage{xcolor}
\usepackage{longtable}
\usepackage{booktabs}

\usepackage{microtype}

\aclfinalcopy 

\title{A BERT-based Distractor Generation Scheme with Multi-tasking and Negative Answer Training Strategies}

\author{ 
  Ho-Lam Chung$^1$, Ying-Hong Chan$^2$, Yao-Chung Fan$^3$ \\
Department of Computer Science and Engineering \\
  National Chung Hsing University, \\
     Taichung, Taiwan \\
  {\tt $^1$holam.chung@protonmail.com}\\
  {\tt $^2$harry831120@gmail.com}\\
  {\tt $^3$yfan@nchu.edu.tw}\\
  }

\date{}

\begin{document}
\maketitle
\begin{abstract}
In this paper, we investigate the following two limitations for the existing distractor generation (DG) methods. First, the quality of the existing DG methods are still far from practical use. There are still room for DG quality improvement. Second, the existing DG designs are mainly for single distractor generation. However, for practical MCQ preparation, multiple distractors are desired. Aiming at these goals, in this paper, we present a new distractor generation scheme with multi-tasking and negative answer training strategies for effectively generating \textit{multiple} distractors. The experimental results show that (1) our model advances the state-of-the-art result from 28.65 to 39.81 (BLEU 1 score) and (2) the generated multiple distractors are diverse and shows strong distracting power for multiple choice question.
\end{abstract}

\section{Introduction}\label{sec:intro}
Given a passage, a question, and an answer phrase, the goal of distractor generation (DG) is to generate context-related wrong options (i.e., distractor) for multiple-choice questions (MCQ). Pioneering research \cite{gao2019generating,yeung2019difficulty,zhou2019coattention} have demonstrated the feasibility of generating distractors based on deep learning techniques. 

\begin{table}[t]
\resizebox{\linewidth}{!}{
\begin{tabular}{l}
\textbf{Example 1} \\ \hline
\multicolumn{1}{|l|}{\textbf{Context} Omitted. (See Appendix)} \\
\multicolumn{1}{|l|}{\textbf{Question}} \\
\multicolumn{1}{|l|}{$\cdot$ Why did Mr.King want to send Henry away?} \\
\multicolumn{1}{|l|}{\textbf{Answer}} \\
\multicolumn{1}{|l|}{\begin{tabular}[c]{@{}l@{}} $\cdot$ Because Henry was too lazy.\end{tabular}} \\
\multicolumn{1}{|l|}{\textbf{Gen. Distractors}} \\
\multicolumn{1}{|l|}{$\cdot d_1:$ Because Henry didn't want to go.} \\
\multicolumn{1}{|l|}{$\cdot d_2:$ Because Henry didn't want to go to the bookstore.} \\ \hline
 \\

\textbf{Example 2} \\ \hline
\multicolumn{1}{|l|}{\textbf{Context} Omitted. (See Appendix)} \\
\multicolumn{1}{|l|}{\textbf{Question}} \\
\multicolumn{1}{|l|}{$\cdot$ Which of the following women would look most attractive?} \\
\multicolumn{1}{|l|}{\textbf{Answer}} \\
\multicolumn{1}{|l|}{$\cdot$ A short red-haired woman who wears a purple hat.} \\
\multicolumn{1}{|l|}{\textbf{Gen. Distractors}} \\
\multicolumn{1}{|l|}{$\cdot d_1:$ A young woman who wears a white hat.} \\
\multicolumn{1}{|l|}{\begin{tabular}[c]{@{}l@{}}$\cdot d_2:$ A woman who wears a white hat.\end{tabular}} \\ \hline
\end{tabular}
}
\caption{Two examples for showing the issue of generating multiple distractors by a simple beam search: Note that the generated distractors (i.e., $d_1$ and $d_2$) are the same statements with only slight word usage difference. Such results lower the distracting power for MCQ preparation.}
\label{tab:my-table-mdg}
\end{table}

While significant advances for DG were reported in the literature, we find that the existing DG results are still far from practical use. In this paper, we investigate the following two issues for distractor generation: (1) \textit{DG quality improvement} and (2) \textit{Multiple distractor generation}. 

\noindent\textbf{DG Quality Improvement}
There is still room to be improved for high-quality distractor generation. 
By manually examining the DG results generated by the existing method, we find that the results are still far from ideal for practical use. Thus, one goal of our research is to improve the DG quality further. 

For the quality issues, in this paper, we explore BERT model's employment for performance improvement. As known, employing transformer-based language
models has shown to be useful for improving
NLP tasks. Thus, we investigate the BERT model's application for DG and report our design in this paper.

\noindent\textbf{Multiple Distractor Generation} The existing DG methods mainly focus on \textit{single} distractor generation. However, for practical MCQ preparation, multiple distractors are desired. For more than one distractor, the existing practice is to keep multiple results given by a beam search strategy. However, we find that in many cases, the generated distractors are all referred to the same concept/thing. In fact, the generated distractors are all from the same latent representation, which brings concerns that they might be semantically similar. In Table \ref{tab:my-table-mdg}, we show two DG examples for this problems. In the illustrated examples, one can observe that the generated distractors are the same statements with only a slight word usage difference. Such results lower the distracting power for MCQ preparation. 

For this limitation, we propose to view multiple distractor generation/selection problems as a \textit{coverage} problem, rather than individually selecting top-\textit{k} distractors based on prediction probability. In other words, we propose to choose a distractor set, which maximizes the difficulty of multiple-choice questions, rather than individually picking results with the highest probability but with similar semantic.


The contributions of this paper are (1) a new DG model based on the BERT model employment. The experiment evaluation with benchmarking datasets shows that our model outperforms the existing best models \cite{zhou2019coattention} and pushes the state-of-the-art result from 28.65 to 39.81 (BLEU 1 score). (2) An investigation to employ the use of multiple-choice question answering task to evaluate the DG performance. (3) An investigation for considering the multiple distractors generation problem as a coverage problem. The experiment result demonstrates that the generated multiple distractors are diverse and show strong distracting power for multiple-choice questions.   

    
    
    
    

The rest of this paper is organized as follows. In Section \ref{sec:BDG}, we introduce our model design for a single distractor generation. In Section \ref{sec:MDG}, we introduce to our multiple distractor schemes and the incorporation of the question-answer models for distractor selection. In Section \ref{sec:exp}, we report the result of performance analysis. In Section \ref{sec:related}, we review the literature related to this work. Finally, Section \ref{sec:conclusion} concludes our study and discusses future works. 
\section{BERT Distractor Generation}\label{sec:BDG}

\subsection{BERT Model Review}
The BERT model and its family \cite{liu2019roberta,lan2019albert} are composed of a stack of multi-layer bidirectional Transformer encoders. The input to a BERT model is a sequence of tokens. For a given token, its input representation to the BERT model is first constructed by summing the corresponding token, segment, and position embeddings. After the input representation, the input embeddings travel through the pre-trained/fine-tuned BERT for task learning and prediction. In general, BERT can be employed in two-level language modeling tasks: sequence-level classification and token-level prediction tasks. For the tasks, there are three special tokens, \texttt{[C]}, \texttt{[S]}, and \texttt{[M]}. The embedding of the \texttt{[C]} token is designed to be used as the aggregate sequence representation for classification tasks. The \texttt{[S]} is designed to distinguish different sentences of a token sequence (to provide/signal information from multiple sentences, as the input token sequence can be a pack of multiple sentences). On the other hand, the \texttt{[M]} token is designed to be used in token-level prediction (e.g., predicting a masked token based on context words or predicting the starting/ending probabilities for span-based tasks such as QA tasks).

As reported in \cite{chan2019recurrent, dong2019unified}, BERT essentially is an auto-encoder language modeling design, which aims to reconstruct the original data from corrupted inputs. If BERT is asked to predict a sequence of consecutive masked tokens, it often produces incoherent and ramble results. For example, when using BERT to predict three consecutive $\mathrm{\texttt{[M]}}\mathrm{\texttt{[M]}}\mathrm{\texttt{[M]}}$ masked tokens, the same prediction result for the tokens are often observed. This is because the context (the information for predicting the tokens) for the masked tokens are nearly the same except for the position embedding, making the generated sentences incoherent. Thus, we take into consideration the previous decoded results for decoding the next distractor token, as will be introduced in the next subsection.

\begin{table*}
\centering
\begin{tabular}{|l|l|l|}
\hline
Iter. & Input Sequence & Predict \\ \hline
1 & \texttt{[C] $C$ [S][M]} & Because  \\ \hline
2 & \texttt{[C] $C$ [S]} Because  \texttt{[M]} & Henry \\ \hline
3 & \texttt{[C] $C$ [S]} Because Henry \texttt{[M]} & didn't \\ \hline
4 & \texttt{[C] $C$ [S]} Because Henry didn't \texttt{[M]} & want \\ \hline
5 & \texttt{[C] $C$ [S]} Because Henry didn't want \texttt{[M]} & to \\ \hline
6 & \texttt{[C] $C$ [S]} Because Henry didn't want to  \texttt{[M]} & go \\ \hline
7 & \texttt{[C] $C$ [S]} Because Henry didn't want to go  \texttt{[M]} & . \\ \hline
8 & \texttt{[C] $C$ [S]} Because Henry didn't want to go.\texttt{[M]} & \texttt{{[}S{]}} \\ \hline
\end{tabular}
\caption{A Running Example for the BDG scheme}
\label{tab:dg}
\end{table*}


\subsection{BERT-based Distractor Generation (BDG)}\label{sec:BDG_architecture}

In a distractor generation scenario, there are three given inputs: (1) a paragraph $P$, (2) an answer $A$, and (3) a question $Q$. For ease of discussion, let $C$ (referred to as a context sequence) denote the sequence of tokens given by concatenating $P$, $Q$, and $A$. 

Our BDG model generates distractor tokens in an auto-regressive manner. Specifically, the BDG model predicts a token at a time based on (1) the given context sequence $C$ and (2) the previously predicted distractor tokens. The BDG model takes multiple iterations to generate a distractor. In Table \ref{tab:dg}, we show a running example of the BDG model. Note that our model predicts a token based on $C$ and the previously generated tokens at each iteration. For example, at Iteration 1, we generate "Because" based on $C$. At Iteration 2, we generate "Henry" based on $C$ and "Because" tokens, and Iteration 3, we generate "didn't" based on $C$, "Because", and "Henry". The generation terminates when \texttt{[S]} is predicted. In this example, "Because Henry didn't want to go." is the final generated result. 

Specifically, the input sequence $X_{i}$ at Iteration $i$ to BERT is 
\[ \begin{split}
X_{i} = (\mathrm{\texttt{[C]}}, C, \mathrm{\texttt{[S]}},\hat{d_{1}}, ...,\hat{d_{i}},\mathrm{\texttt{[M]}})
\end{split} \]
Let $\mathbf{h}_{\texttt{[M]}} \in \mathbb{R}^h$ denote the hidden representation of \texttt{[M]} of $X_{i}$ returned by BERT transformer stacks. The prediction of $\hat{d_i}$ is given by a linear layer transformation $\mathbf{W}_{\texttt{DG}} \in \mathbb{R}^{h\times |V|}$ and a softmax activation to all vocabulary dimension as follows.  
$$p(w|X_i) = softmax(\mathbf{h}_{\texttt{[M]}}\cdot\mathbf{W}_{\texttt{DG}}+\mathbf{b}_{\texttt{DG}})$$ 
$$\hat{d_{i+1}}= \mathrm{argmax}_w {Pr(w|X_i)} $$



Subsequently, the newly generated token $\hat{d_i}$ is appended into $X_{i+1}$ and the distractor generation process is repeated based on the new $X_{i+1}$ until \texttt{[S]} is predicted. Our loss function is as follows.
$$\underset{\theta}{\text{minimize}}-\sum_{\forall (C,  D)} \sum_{i=0}^{|D|}(\text{log}_2p(d_{i+1}|{C, d_{1:i};\theta}))$$


\subsection{Multi-task with Parallel MLM}

From the experiment results (will be presented in the later section), we see the BDG model advances the state-of-the-art result \cite{zhou2019coattention} from 28.65 to 35.30 (BLEU 1 score). While the token-level evaluation result looks promising, we find that generation results still have room to be improved.

For performance improvement, we first propose to jointly train BDG and a parallel MLM (P-MLM) architecture for distractor generation to enhance the quality of BDG. The P-MLM scheme for generating distractors is structured as follows. 

For a given context $C$, the input sequence $X$ to P-MLM model is formulated as
\[ \begin{split}
X = (\mathrm{\texttt{[C]}}, C, \mathrm{\texttt{[S]}}, \mathrm{[\texttt{M}]_{d_1}}, \mathrm{[\texttt{M}]_{d_2}}, ...,\mathrm{[\texttt{M}]_{d_{|D|}}})
\end{split} \]
Let $\mathbf{h}_{\mathrm{\texttt{[M]}_{d_i}}} \in \mathbb{R}^h$ denote the hidden representation of $\mathrm{\texttt{[M]}_{d_i}}$ of $X$ returned by BERT transformer stacks. The prediction of $\hat{q_i}$ is given by a linear layer transformation $\mathbf{W}_{\texttt{P-MLM}} \in \mathbb{R}^{h\times |V|}$ and applying a softmax activation to all vocabulary dimension as follows.  
$$p(w|X) = softmax(\mathbf{h}_{\mathrm{[\texttt{M}]_{d_i}}}\cdot\mathbf{W}_{\texttt{P-MLM}}+\mathbf{b}_{\texttt{P-MLM}})$$ 
$$\hat{d_{i}}= \mathrm{argmax}_w {Pr(w|X)} $$
The loss function for P-MLM is 
$$\underset{\theta}{\text{minimize}}-\sum_{\forall (C,  D)}\phi_{_{\texttt{P-MLM}}}(C,D)$$
$$\phi_{_{\texttt{P-MLM}}}(C,D)=\sum_{\forall d_i}\text{log}_2p(d_i|{C, \mathrm{[\texttt{M}]}_{d_i};\theta})$$

We propose to jointly train P-MLM and BDG by the following multi-tasking loss function. Note that $\gamma$ is a hyper-parameter controlling the weighting between the two tasks. See also the effect of the $\gamma$ value in Subsection \ref{subsec:parameter_study}. 


$$
\underset{\theta}{\text{minimize}}-\sum_{\forall (C, D)}[\phi_{_{\texttt{BDG}}}(C,D)+\gamma\cdot\phi_{_{\texttt{P-MLM}}}(C,D)]
,$$
$$
    \phi_{_{\texttt{BDG}}}(C,D)=\sum_{i=0}^{|D|}(\text{log}_2p(d_{i+1}|{C, d_{1:i};\theta}))
$$

The multi-task design is motivated by the following observations. First, as mentioned, we target at learning distractor generation from real reading comprehension examination (RACE-like MCQ), and we find that many questions in the RACE dataset are summary-oriented; many questions are about "what is the best title for this passage?" or "what is this passage about?" Such questions require the model to have the capability of passage semantic summarization. While the original BDG scheme design successfully generates fluent question sentences, we find that it may over-fit in sentence writing and under-fit in learning the passage semantic understanding capability. Note that the sequential-MLM design (BDG) essentially is a one-by-one masked token prediction architecture. Such a method may over-focus on the guess of a single token and ignore the overall semantic understanding. Thus, we propose to incorporate the multi-task learning setting to prevent the potential over-fitting problem. From the experiments, we find the multi-task learning setting indeed improves the quality of distractor generation.

\subsection{Answer Negative Regularization}
In addition to the multi-task design, from the DG result examination, we find another observation that in many cases, there is an \textit{answer copying problem}; the generated distractors are similar to the given answers. To better see this phenomenon, we experiment to count such cases. In the following table, we show the number of cases that the generated distractor $\hat{D}$ has a token-level similarity score greater than $0.95$ with respect to the answer $A$. We also show the cases for the gold distractors (the human-invented distractors from the RACE dataset). By comparison in Table \ref{tab:ARPM}, there is a significant gap between the human-invented distractors and the model generated ones. 

\begin{table}
\centering
    \begin{tabular}{|l|l|l|}
    \hline
     & P.M. & Gold \\ 
    \hline
    \# of cases on BLEU 1   & 57 & 12 \\ \hline
    \# of cases on BLEU 2   & 55 & 4 \\ \hline
    \# of cases on BLEU 3   & 48 & 0 \\ \hline
    \# of cases on BLEU 4   & 35 & 0 \\ \hline
    \# of cases on ROUGE-L   & 55 & 1 \\\hline
    \end{tabular}
    \caption{Answer Copying Problem on P.M.}
    \label{tab:ARPM}
\end{table}

Motivated by the answer copying problem, we propose to incorporate a loss (referred to as \textit{answer negative loss}) to discourage predicting tokens in $A$ when predicting $\hat{d_i}$. With the answer negative loss, our loss function for BDG is as follows.

\begin{equation}
\begin{split}
&\underset{\theta}{\text{minimize}}-\sum_{\forall (C, D)}(\phi_{_{\texttt{AN}}}(C,D)+\gamma\cdot\phi_{_{\texttt{P-MLM}}}(C,D)), \\
& \begin{split}
    \phi_{_{\texttt{AN}}}=&\sum_{i=0}^{|D|}(\text{log}_2p(d_{i+1}|{C,d_{1:i};\theta})+\\&\sum_{\forall a_j\in A}\text{log}_2(1-p(a_{j}|{C,\mathrm{[\texttt{M}]}_{a_j};\theta}))  
    \end{split}
\end{split}
  \label{eq:SLUL}
\end{equation}





The design of answer negative loss is motivated by that we expect to regulate the generated distractor $\hat{D}$ to use words different from $A$.

\begin{figure}[t]
    \centering
    \includegraphics[width=1.\columnwidth]{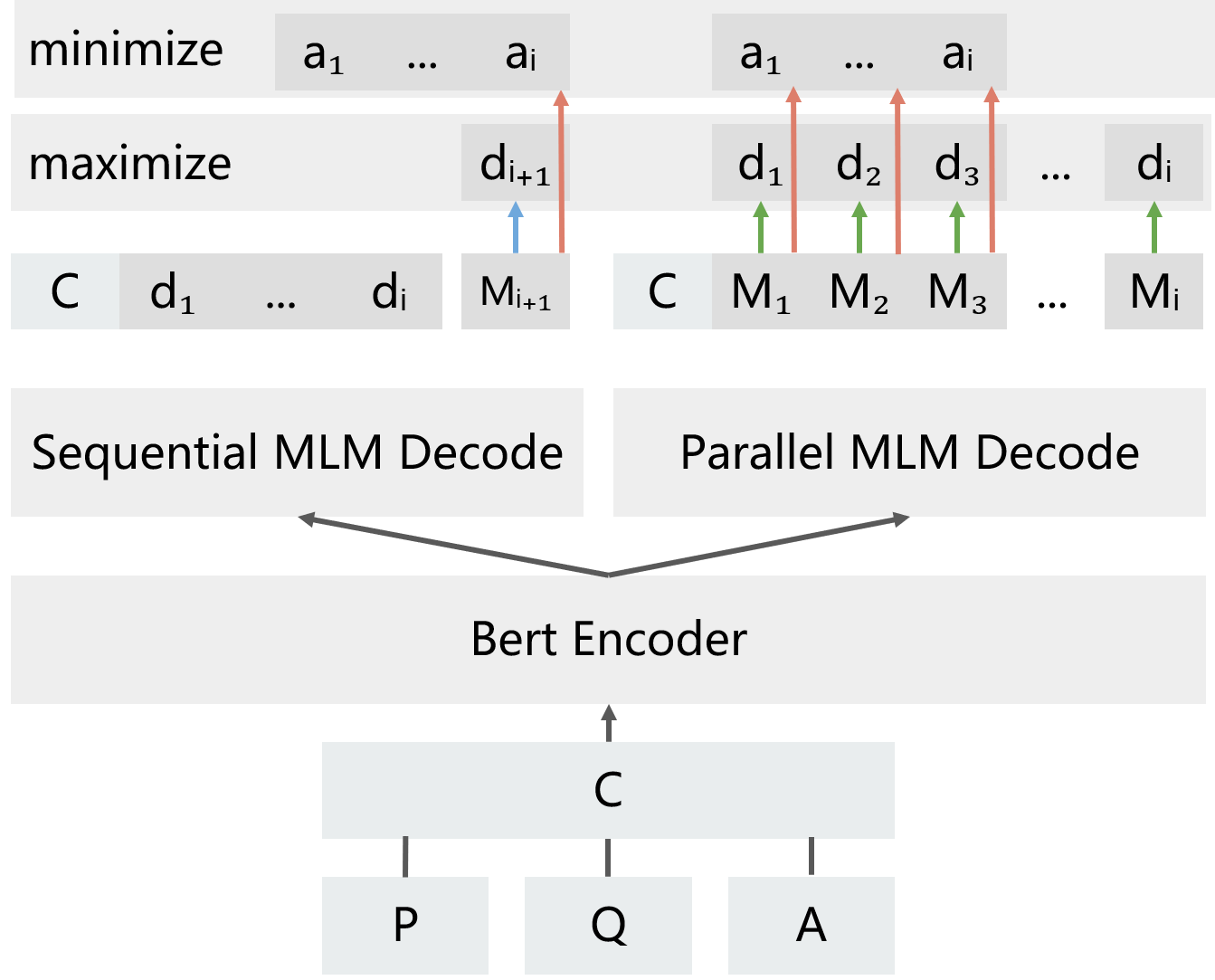}
    \caption{The Multi-tasking Architecture}
    \label{fig:Multi_task}
\end{figure}

The overall architecture for training our BDG model is shown in Figure \ref{fig:Multi_task}. The core structure for our distractor generation is mainly based on the sequential recurrent MLM decoding mechanism. That is, during the the testing stage, we use the results from the sequential recurrent MLM decoding part. However, during the training stage, we incorporate the parallel MLM decoding mechanism by jointly considering answer negative regularization and sentence-level distractor loss, as shown in the right-part of the architecture in Figure \ref{fig:Multi_task}.

\section{Multiple Distractor Generation}\label{sec:MDG}
\subsection{Selecting Distractors by Entropy Maximization}\label{sec:EM}
As mentioned, another point that can be improved for DG is that the existing methods mainly focus on single distractor generation. For having more than one distractor, the existing practices are to select the results on different beam search paths as multiple options for distractor generation, which lowers the power of distracting a reader for MCQ preparation. 

Our viewpoint is to select a distractor set (by considering semantic diversity) rather than individually selecting top-k distractors based on prediction probability. 

Based on this view, we propose to incorporate a multi-choice reading comprehension (MRC) model for ranking/selecting distractor sets. First, let $\mathbb{M}_{\texttt{MRC}}$ be a MRC model. Note that $\mathbb{M}_{\texttt{MRC}}$ takes a passage $P$, a question $Q$, and a set of options (including an answer $A$ and distractors $D_1, D_2, ..., D_n$) as input and outputs [$p_A, p_{D_1}, ...,p_{D_n}$] as the answer probabilities of the options. $\mathbb{M}_{\texttt{MRC}}$ is trained by maximizing the answer probability $p_A$ while minimizing the probabilities [$p_{D_1}, ...,p_{D_n}$]. 

With $\mathbb{M}_{\texttt{MRC}}$, our idea is as follows. First, let $\mathbb{DG}_{\texttt{BDG}}$ be a BDG model for distractor generation. Also, let $\hat{D}=\{\hat{d}_1, \hat{d}_2, ..., \hat{d}_n\}$ be the set of generated distractors by the BDG model. In a common MCQ setting, there are four options (one answer $A$ and three distractors $d_i, d_j, d_k$) for each question. Our idea is to enumerate all possible triples from $\{\hat{d}_1, \hat{d}_2, ..., \hat{d}_n\}$. That is, we have a triple set 
\[
  \{ (d_i, d_j, d_k)| i\neq j\neq k, d_i, d_j, d_k \in \hat{D} \}
\]

For a given passage $P$, question $Q$, and answer $A$, our goal is to find a triple ($d_i, d_j, d_k$) to form an option set $O$ (i.e., \{$d_i, d_j, d_k$, $A$\} ) that maximizes the following entropy function.

\begin{equation}\label{eq:entropy}
    \underset{}{\text{maximize}}-\sum_{\forall o_i\in O}p_{o_i}log_2p_{o_i}
\end{equation}

\subsection{BDG-EM}\label{sec:model-ensemble}
The idea of selecting distractors by entropy maximization can be further generalized by employing multiple DG models. For having multiple DG models, our idea is to leverage the variants of the BDG model (i.e., models with/without answer negative regularization or with/without both answer negative regularization and P-MLM multi-task training). Let $\hat{D}, \hat{D}_{\texttt{PM}}$, and $\hat{D}_{\texttt{PM+AN}}$ be the BDG model without both answer negative regularization and P-MLM multi-task training, the BDG model without answer negative regularization, and the full BDG model. That is, we have a triple set as follows.
\[
  \{ (d_i, d_j, d_k)| d_i \in \hat{D}, d_j \in \hat{D}_{\texttt{PM}}, d_k \in \hat{D}_{\texttt{PM+AN}} \}
\]

With the triple set, the set that maximizes Eq. (\ref{eq:entropy}) is selected as final distractors.

\section{Performance Evaluation}\label{sec:exp}

\subsection{Experimental Settings}

\noindent\textbf{Datasets}
We follow the setting \cite{gao2019generating} to evaluate our framework with the RACE \cite{lai2017large} dataset.
RACE contains 27,933 articles with 97,687 questions from English examinations of Chinese students from grade 7 to 12. We use data split setting from \cite{gao2019generating}. Table \ref{table:training data stat} reports the statistics for the test data set. All sentences are tokenized by the WordPiece tokenizer \cite{wu2016google}.

\noindent\textbf{Implementation Details}
Our models are implemented based on huggingface transformers framework \cite{Wolf2019HuggingFacesTS}. All experiments are based on bert-base-cased model. For optimization in the training, we use AdamW as the optimizer and the initial learning rate 5e-5 for all baselines and our model. The maximum number of epoch is set to 6 with a batch size of 30 on two RTX Titan GPUs. We also make our code and model available at \href{https://github.com/voidful/BDG}{https://github.com/voidful/BDG} 

\subsection{Compared Methods}
In the experiments, we mainly compare the following distractor generation methods. 
\begin{itemize}
     
    \item \textbf{CO-Att.} We compare with the state-of-the-art method reported in \cite{zhou2019coattention}. The model is based on LSTM augmented by co-attention mechanism. 
    \item \textbf{DS-Att.} We also compare with the method based on LSTM augmented by dynamic and static attention designed reported in \cite{gao2019generating}. This method is served as a baseline for distractor generation based on seq2seq RNN architectures. 
    \item \textbf{GPT} We also experiment with a model based on GPT \cite{radford2018improving} to learn the distractor generation. This scheme can be served as a baseline based on transformer-based pre-trained model. 
    \item \textbf{BDG} The scheme without the answer negative technique and parallel masked-LM multi-task training.
    \item \textbf{BDG$_\texttt{PM}$} The BDG scheme with the parallel masked-LM multi-task training ($\gamma=1$).
    \item \textbf{BDG$_\texttt{AN+PM}$} The BDG scheme with both techniques ($\gamma=1$).
\end{itemize}

\begin{table}
  \centering
  \begin{tabular}{ll}
    \midrule
    Train samples    &  96501     \\
    Test  samples     &  12284    \\
    \midrule
    Avg.article length  &  335.6     \\
    Avg.distractor length   &  8.6   \\
    Avg.question length &  10.0        \\
    Avg.answer length   & 8.3        \\
    \hline
    Avg.distractor number &  2.1 \\
    \hline
  \end{tabular}
  \caption{Training Data Statistics}
  \label{table:training data stat}
\end{table}

\begin{table*}
    \centering
    \begin{tabular}{|l|l|l|l|l|l|}
    \hline
                          & BLEU 1         & BLEU 2         & BLEU 3         & BLEU 4        & ROUGE L         \\ 
    \hline
    BDG$_\texttt{AN+PM}$  & 39.52    & 24.29 & 17.28 & 13.28 & 33.40           \\ 
    \hline
    BDG$_\texttt{PM}$     & \textbf{39.81}          & \textbf{24.81}          & \textbf{17.66}          & \textbf{13.56}          & \textbf{34.01}  \\ 
    \hline
    BDG                   & 35.30          & 20.65          & 13.66          & 9.53          & 31.11           \\ 
    \hline
    GPT                   & 36.49          & 20.75          & 13.31          & 9.31          & 31.59           \\ 
    \hline
    DS-Att.               & 27.32          & 14.69          & 9.29           & 6.47          & 15.12           \\ 
    \hline
    CO-Att.               & 28.65          & 15.15          & 9.77           & 7.01          & 15.39           \\
    \hline
    \end{tabular}
    \caption{Performance Comparison on Token Scores}
    \label{tab:toke_score_overview}
\end{table*}

\begin{table*}
    \centering
    \begin{tabular}{|l|l|l|l|l|l|l|l|} 
\hline
               & BDG$_\texttt{AN+PM}$  & BDG$_\texttt{PM}$    & BDG & GPT & Gold & Random  \\ 
\hline
BLEU 1   & \textbf{43}           & 57                        & 115 & 124 & 12   & 0       \\ 
\hline
BLEU 2   & \textbf{40}           & 55                         & 115 & 121 & 4   & 0       \\ 
\hline
BLEU 3   & \textbf{37}           & 48                        & 109 & 109 & 0    & 0       \\ 
\hline
BLEU 4   & \textbf{30}           & 35                        & 97  & 88  & 0    & 0       \\ 
\hline
ROUGE-L & \textbf{42}           & 55                      & 122 & 123 & 1    & 0       \\
\hline
\end{tabular}
    \caption{The Effect on Mitigating Answer Copying Problem}
\label{tab:repeating-problem}
\end{table*}

\subsection{Token Score Comparison}
We employ BLEU score \cite{papineni2002bleu} and ROUGE (L) \cite{lin2004rouge} scores to evaluate the performance of the compared methods. The BLEU scores evaluate average n-gram precision on a set of reference sentences, with a penalty for overly long sentences. The ROUGE (L) measure is the recall of longest common sub-sequences.

The comparison results are summarized in Table \ref{tab:toke_score_overview}. There are three observations to note. First, one can see that our models significantly outperform the existing methods (i.e., DS-Att. and CO-Att.). Our best performing model advances the state-of-the-art result from 28.65 to 39.81 (BLEU 1 score). Second, as shown, the methods based on transformer models outperform the RNN-based models. This result again demonstrates the effectiveness of the employment of pre-trained transformer model to the downstream tasks. Third, one may notice that our models based on BERT outperforms the GPT-based model. We believe the reason is that the distractors in the RACE data set is mostly a summary type sentence that requires semantic understanding. The GPT-based model may over-focus on sentence writing, and fail to capture the whole context to generate summary-type sentences, and therefore obtain lower scores. 

We also provide experiment results to observe the effectiveness on reducing the answer copying problem discussed in Subsection \ref{sec:BDG}. In Table \ref{tab:repeating-problem}, we show the number of cases that the generated distractor $\hat{D}$ has a token-level similarity score greater than $0.95$ with respect to the context answer $A$. From the experiment result, we see that there are significant improvement made by the BDG schemes.


\subsection{MCQ Model Accuracy Comparison}
In this set of experiment, we evaluate the DG quality by the RACE reading comprehension task \cite{lai2017large}. Our idea is that a poorly generated DG result will reduce the difficulty of a MCQ task. Thus, we propose to incorporate a MCQ answering model (also trained by the RACE dataset) to evaluate the accuracy of a multiple-choice question with the distractors generated by the compared model. Specifically, given $C$, $Q$, and $A$, we generate three distractors $D_1$, $D_2$, and $D_3$, and then submit the multiple-choice question to the RACE model. Randomly generated results will be the easiest task to solve, and the best generated results will bring challenges to the MCQ model. Therefore, we use the accuracy of the model as a metric. The higher the accuracy, the worse the generation quality.

The training details of the RACE model is as follows. We use PyTorch  Transformers\cite{Wolf2019HuggingFacesTS} and the roberta-base-openai-detector fine-tuned by OpenAI \cite{solaiman2019release} with max 512 tokens to implement the model. AdamW with a Learning rate = 1e-5 is used for fine-tuning. The model is trained for 10 epoch on 2 GPUs (V100) with gradient accumulation per two steps, which makes the batch size approximately equal to 18. Model checkpoints are saved and evaluated on the validation set every 5,000 steps. We select the top checkpoint based on evaluation loss on the validations set. The RACE dataset includes middle and high dataset. The total number of passages and questions is 27,933 and 97,687 respectively. Middle dataset averages about 250 words per passage while the High dataset averages 350 words per passage.

\begin{table}
    \centering
    \begin{tabular}{|l|r|}
        \hline
         & \multicolumn{1}{l|}{Accuracy} \\ \hline
        Random Selected Distractors & 88.10\% \\ \hline
        Gold Distractor & 78.00\% \\ \hline
        GPT & 78.07\% \\ \hline
        BDG & 73.96\% \\ \hline
        BDG$_\texttt{PM}$ & 74.34\% \\ \hline
        BDG$_\texttt{AN+PM}$ & 74.05\% \\ \hline
        BDG$_\texttt{EM}$  & \textbf{69.44\%} \\ \hline
    \end{tabular}
\caption{Comparison by MCQ Accuracy}
\label{tab:rc result}
\end{table}

In this set of experiment, we compare BDG, BDG$_\texttt{PM}$, BDG$_\texttt{AN+PM}$, the BDG model with entropy maximization (called BDG$_\texttt{EM}$) (introduced in Subsection \ref{sec:model-ensemble}) by setting the beam search size to 3, and the BDG model ensemble introduced in Subsection \ref{sec:model-ensemble}. In addition, we also experiment with the GPT, a scheme that takes randomly selected distractors from the data as the DG result, and the scheme uses the gold distractors. The results of the compared methods are summarized in Table \ref{tab:rc result}. 

We have the following findings to note about the results shown in Table \ref{tab:rc result}. First, as expected, the method with randomly selected distractors makes the MCQA model has the highest accuracy, as the randomly selected distractors obviously lower the difficulty of MCQ task. Second, all our models outperform the MCQ with the gold distractors, showing the effectiveness of the proposed models. Third, as expected, our BDG$_{\texttt{EM}}$ provides the best performing result on this metric.

\subsection{Qualitative Examination by Case Study}
In this subsection, we present showcases to see the improvement on multiple distractor generation scenario. We use the same examples introduced in Section \ref{sec:intro} for comparison. First, as mentioned, the naive employment of beam search strategy produces similar DG results. As shown in the examples, the distractors generated by BDG are about the same concept. However, as shown in Table \ref{tab:qualitative_Examination}, we see the BDG$_{\texttt{EM}}$ produce more diverse distractors with respect to each other. The results demonstrate the effectiveness of our BDG$_{\texttt{EM}}$ scheme for generating multiple distractors for MCQ preparation.

\begin{table}[t]
\resizebox{\linewidth}{!}{
\begin{tabular}{l}
\textbf{Example 1} \\ \hline
\multicolumn{1}{|l|}{\textbf{Context} Omitted. (See Appendix)} \\
\multicolumn{1}{|l|}{\textbf{Question}} \\
\multicolumn{1}{|l|}{$\cdot$ Why did Mr.King want to send Henry away?} \\
\multicolumn{1}{|l|}{\textbf{Answer}} \\
\multicolumn{1}{|l|}{\begin{tabular}[c]{@{}l@{}} $\cdot$ Because Henry was too lazy.\end{tabular}} \\\hline
\multicolumn{1}{|l|}{\textbf{BDG}} \\
\multicolumn{1}{|l|}{$\cdot d_1:$ Because Henry didn't want to go.} \\
\multicolumn{1}{|l|}{$\cdot d_2:$ Because Henry didn't want to go to the bookstore.} \\ 
\multicolumn{1}{|l|}{$\cdot d_3:$ Because Henry didn't want to go out.} \\ 
\hline
\multicolumn{1}{|l|}{\textbf{BDG$_{\mathbf{EM}}$}} \\
\multicolumn{1}{|l|}{$\cdot d_1:$ Because Henry didn't want to go.} \\
\multicolumn{1}{|l|}{$\cdot d_2:$ Because Henry wanted to be rich.} \\ 
\multicolumn{1}{|l|}{$\cdot d_3:$ Because Henry wanted to be a clever man.} \\

\hline
 \\

\textbf{Example 2} \\ \hline
\multicolumn{1}{|l|}{\textbf{Context} Omitted. (See Appendix)} \\
\multicolumn{1}{|l|}{\textbf{Question}} \\
\multicolumn{1}{|l|}{$\cdot$ Which of the following women would look most attractive?} \\
\multicolumn{1}{|l|}{\textbf{Answer}} \\
\multicolumn{1}{|l|}{$\cdot$ A short red-haired woman who wears a purple hat.} \\\hline
\multicolumn{1}{|l|}{\textbf{BDG}} \\
\multicolumn{1}{|l|}{$\cdot d_1:$ A young woman who wears a white hat.} \\
\multicolumn{1}{|l|}{\begin{tabular}[c]{@{}l@{}}$\cdot d_2:$ A woman who wears a white hat .\end{tabular}} \\ 

\hline
\multicolumn{1}{|l|}{\textbf{BDG$_{\mathbf{EM}}$}} \\
\multicolumn{1}{|l|}{$\cdot d_1:$ A short black woman with big, round faces.} \\
\multicolumn{1}{|l|}{$\cdot d_2:$ A young woman who doesn’t like a white hat.} \\ 
\multicolumn{1}{|l|}{$\cdot d_3:$ A little woman who wears a pink hat.} \\

\hline
\end{tabular}
}
\caption{Qualitative Examination by Case Study}
\label{tab:qualitative_Examination}
\end{table}

\begin{table}
\resizebox{\linewidth}{!}{
    \centering
    \begin{tabular}{|l|l|l|l|l|l|}
    \hline
                          & BLEU 1         & BLEU 2         & BLEU 3         & BLEU 4        & ROUGE L         \\ 
    \hline
    PM($\gamma$=1)  & 36.97    & 22.07 & 14.82 & 10.50 & 32.64           \\ 
    \hline
    PM($\gamma$=2)  & 38.45          & 23.21          & 15.81          & 11.36          & 33.18  \\ 
    \hline
    PM($\gamma$=3)  & 39.23          & 24.27          & 17.04          & 12.78          & 33.82           \\ 
    \hline
    PM($\gamma$=4)  & 39.22          & 24.24          & 17.08          & 12.95          & 34.05           \\ 
    \hline
    PM($\gamma$=5)  & 39.74          & 24.50          & 17.29           & 13.09          & \textbf{34.11}           \\ 
    \hline
    PM($\gamma$=6)  & \textbf{39.81}          & \textbf{24.81}          & \textbf{17.66}          & \textbf{13.56}          & 34.01           \\ 
    \hline
    PM($\gamma$=7)  & 39.37          & 24.13          & 17.09           & 13.07          & 33.45           \\ 
    \hline
    AN+PM($\gamma$=1)  & 37.49    & 22.08 & 13.73 & 10.44 & 32.40           \\ 
    \hline
    AN+PM($\gamma$=2)  & 38.25          & 22.81          & 15.33          & 10.91          & 32.99  \\ 
    \hline
    AN+PM($\gamma$=3)  & 38.71          & 23.54          & 16.26          & 12.04          & \textbf{33.82}           \\ 
    \hline
    AN+PM($\gamma$=4)  & 38.84          & 23.70          & 16.57          & 12.46          & 33.53           \\ 
    \hline
    AN+PM($\gamma$=5)  & 39.19          &23.97          & 16.96          & 12.92          & 33.67           \\ 
    \hline
    AN+PM($\gamma$=6)  & \textbf{39.58}          & 24.23          & 17.11           & 13.11          & 33.38          \\ 
    \hline
    AN+PM($\gamma$=7)  & 39.52          & \textbf{24.29}          & \textbf{17.28}           & \textbf{13.28}          & 33.40          \\ 
    \hline

    \end{tabular}
    }
    \caption{Performance Comparison on Token Scores with Different $\gamma$ Settings}
    \label{tab:toke_score_pm_overview}
\end{table}

\subsection{Parameter Study on $\gamma$}\label{subsec:parameter_study}
In this subsection, we examine the effects on varying the values of the parameter $\gamma$. In Table \ref{tab:toke_score_pm_overview}, we show the results. From the result, we can see that the best setting for $\gamma$ is 6, and for BDG trained by answer negative and parallel-MLM, the best setting for $\gamma$ is 7.

\section{Related Work}\label{sec:related}
The DG research can be categorized from different perspectives. First, for DG task type, there are two main task categories for DG: cloze-style distractor generation and reading comprehension (RC) distractor generation. In cloze-style DG task, it is viewed as a word filling problem. In general, the first step is to extract distractor candidates from context or some knowledge base, and then the next step is to rank the extracted distractors as a final result. Along this direction, the models are mainly based on similarity heuristic \cite{sumita2005measuring,mitkov2006computer,guo2016questimator,ren2020knowledge} or supervised machine learning way \cite{liang2018distractor,yeung2019difficulty}. The distractors generated for cloze-style DG are mainly word/phrase level. On the other hand, the RC-type QG focuses on generating sentence-level distractors for reading comprehension level testing, such as summarizing article or understanding author opinion \cite{gao2019generating,zhou2019coattention}. For the sentence-level distractors, neural models are commonly employed as it is difficult to generate a semantic rich and fluent distractor from question, content, and answer.
In this paper, we also focus on generative sentence-level DG for RC task. However, as mentioned in the introduction, we find the existing DG results are still far from human level. The best SOTA result (in terms of BLEU 1 score) is 29, which is far from the ideal result for practical use. Aiming at this point, we explore the employment of transformer-based pre-trained models for performance improvement. For clarity of comparison, we summarize the existing studies on distractor generation in Table \ref{tab:rel2}.

\begin{table*}[t]
\centering
\resizebox{\textwidth}{!}{
\begin{tabular}{|l|l|l|l|l|l|l|l|} 
\hline
 & \multicolumn{2}{l|}{Distractor Level} & \multicolumn{2}{l|}{Answer Type} & \multicolumn{2}{l|}{Method Type} & Model~ \\ 
\cline{2-8}
 & Word/phrase & Sentence & Cloze & R.C. & Extractive & Generative & Type \\ 
\hline
\citealt{gao2019generating} & Y & Y &  & Y &  & Y & RNN \\ 
\hline
\citealt{zhou2019coattention} & Y & Y &  & Y &  & Y & RNN \\ 
\hline
\citealt{araki2016generating} & Y &  & Y &  & Y &  & Non-neural model \\ 
\hline
\citealt{welbl2017crowdsourcing} & Y &  &  & Y & Y &  & Random forests \\ 
\hline
\citealt{guo2016questimator} & Y &  & Y &  & Y &  & Word2Vec \\ 
\hline
\citealt{kumar2015revup} & Y & Y & Y &  & Y &  & SVM \\ 
\hline
\citealt{liang2017distractor} & Y &  & Y &  &  & Y & GAN \\ 
\hline
\citealt{liang2018distractor} & Y & Y &  & Y & Y &  & Non-neural model \\
\hline
\end{tabular}
}
\caption{An Overview of the Existing DG works}
\label{tab:rel2}
\end{table*}
\section{Conclusion}\label{sec:conclusion}
We present a state-of-the-art neural model based on a pre-trained transformer-based model for DG. We introduce two techniques, Answer Negative Regularization and Multi-task with Parallel MLM, to boost the DG performance. In addition, we also introduce BDG ensemble with an entropy maximization mechanism to enhance the DG quality by leveraging a reading comprehension model. By experimental evaluation, our models outperform the existing best performing models and advances the state-of-the-art result to 39.81 (BLEU 1 score).

\section*{Acknowledgement}
This work was supported by the Ministry of Science and Technology, Taiwan, under projects No. 109-2221-E-005-058-MY3 and 107-2221-E-005-064-MY2
\bibliographystyle{acl_natbib}
\bibliography{reference}
\begin{table*}
\textbf{Appendix}\\
\resizebox{\textwidth}{!}{
    \begin{tabular}{|p{0.15\textwidth}|p{0.85\textwidth}|}
    \hline
    Content & \textit{The building is shaking. A woman with a baby in her arms is trying to open the door, but fails. Finding no way, she rushes into her bedroom and there they survive the earthquake. In a factory building, as the workshop floor swings under the terrible shaking, workers run for safety. Some hide under the machines and survive, but others who try to run outside are killed by the falling ceilings. These scenes, played by actors and actresses, are from a film of science education Making a Split Second Decision shown in 1998 on China Central TV in memory of Tangshan Earthquake. By studying actual cases in the earthquake areas and scientific experiments, experts find that buildings remain untouched for the first 12 seconds of an earthquake. In this short time, one has the best chance of surviving an earthquake by staying near the inside walls, in bedrooms and under beds, experts concluded in the film. "Earthquakes seem to catch the lives of those who run," said many survivors in the earthquake areas, describing how their friends were killed on the doorways or along the stair steps as they tried to get out of the building. Their advice was proved in the film, "Take a hiding-place where you are rather than run, unless you are sure you can reach a safe open place in ten seconds."} \\ \hline
    Question          & The workers who try to run outside the building die because?                  \\ \hline
    Answer            & \textbf{They don't have enough time to run outside.}                \\ \hline
    Distractor        & \textbf{They don't know how to get out of the building.}                \\ \hline
    \end{tabular}
}
    \caption{BDG showcase}
    \label{app:BDG example}
\end{table*}

\begin{table*}
\resizebox{\textwidth}{!}{
    \begin{tabular}{|p{0.15\textwidth}|p{0.85\textwidth}|}
    \hline
    Content & \textit{Henry found work in a bookstore after he finished middle school. He wouldn't do anything but wanted to get rich. Mr.King thought he was too lazy and was going to send him away. Henry was afraid and had to work hard. It was a cold morning. It was snowing and there was thin ice on the streets. Few people went to buy the books and the young man had nothing to do. He hated to read, so he watched the traffic. Suddenly he saw a bag fall off a truck and it landed by the other side of the street. \"It must be full of expensive things.\" Henry said to himself. \"I have to get it, or others will take it away.\" He went out of the shop and ran across the street. A driver saw him and began to whistle, but he didn't hear it and went on running. The man drove aside, hit a big tree and was hurt in the accident. Two weeks later Henry was taken to court. A judge asked if he heard the whistle when he was running across the street. He said that something was wrong with his ears and he could hear nothing. "But you've heard me this time." said the judge. "Oh , I'm sorry. Now I can hear with one ear." "Cover the ear with your hand and listen to me with your deaf one. Well, can you hear me ?" " No, I can't, Sir."} \\ \hline
    Question           & Why did Mr.King want to send Henry away?               \\ \hline
    Answer             & \textbf{Because Henry was too lazy.}                 \\ \hline
    BDG                & Because Henry didn't want to go.                      \\ \cline{2-2}
    \multirow{3}{*}{}  & Because Henry didn't want to go out.                  \\ \cline{2-2}
                       & Because Henry didn't want to go to the bookstore.     \\ \hline
    BDG ensemble       & Because Henry didn't want to go.                   \\ \cline{2-2}
    \multirow{3}{*}{}  & Because Henry wanted to be rich.                      \\ \cline{2-2}
                       & Because Henry wanted to be a clever man.              \\ \hline
    \end{tabular}
}
\caption{Context for Example 1}
\end{table*}

\begin{table*}
\resizebox{\textwidth}{!}{
    \begin{tabular}{|p{0.15\textwidth}|p{0.85\textwidth}|}
    \hline
    Content & \textit{Most of the time, people wear hats to protect themselves from weather conditions . Hats are also worn to show politeness and as signs of social position. But nowadays, hats, especially women's hats, are much more than that. More exactly, hats have changed into fashion and style symbols by many movie stars. What's more, people now consider many different features when choosing even a simple hat. Many designers point out that, when choosing the right hat, it's important to consider the color of your skin as well as your hair, your height, and the shape of your face. First of all, the color of the hat should match the color of your skin and hair. For instance, black hats should be avoided if you are dark skinned. If a purple hat is placed on top of red hair, one will look as attractive as a summer flower. Second, the height of the hat is also an important point. Tall women should not go for hats with tall crowns, just as short women should choose hats with upturned brims to give the look of height. Third, and most importantly, the shape of the face decides the kind of hat one should pick. A small, gentle hat that fits the head looks good on a small face. However, women with big, round faces should choose a different style. As the saying goes, \"Fine feathers make fine birds.\" A good hat can not only help your dress but also support your features, so why not choose the best possible one next time you want to be in public?} \\ \hline
    Question          & According to the article, which of the following women would look most attractive?                  \\ \hline
    Answer            & \textbf{A short red-haired woman who wears a purple hat.}                \\ \hline
    BDG               & A young woman who wears a white hat.                                 \\ \cline{2-2}
    \multirow{3}{*}{} & A young woman who doesn't like a white hat.          \\ \cline{2-2}  
                      & A woman who wears a white hat.                         \\ \hline
    BDG ensemble      & A short black woman with big, round faces.                  \\ \cline{2-2}
    \multirow{3}{*}{} & A young woman who doesn't like a white hat.                         \\ \cline{2-2}  
                      & A little woman who wears a pink hat.                              \\ \hline
                      
    \end{tabular}
}
\caption{Context for Example 2}
\end{table*}

\begin{table*}
\resizebox{\textwidth}{!}{
    \begin{tabular}{|p{0.15\textwidth}|p{0.85\textwidth}|}
    \hline
    Content & \textit{Memory, they say, is a matter of practice and exercise. If you have the wish and really made a conscious effort, then you can quite easily improve your ability to remember things. But even if you are successful, there are times when your memory seems to play tricks on you. Sometimes you remember things that really did not happen. One morning last week, for example, I got up and found that I had left the front door unlocked all night, yet I clearly remember locking it carefully the night before. Memory "trick" work the other way as well. Once in a while you remember not doing something, and then find out that you did. One day last month, for example, I was sitting in a barber shop waiting for my turn to get a haircut, and suddenly I realized that I had got a haircut two days before at the barber shop across the street from my office. We always seem to find something funny and amusing in incidents caused by people's forgetfulness or absent-mindedness. Stories about absent-minded professors have been told for years, and we never got tired of hearing new ones. Unfortunately, however, absent-mindedness is not always funny. There are times when "trick" of our memory can cause us great trouble.} \\ \hline
    Question          & Which of the following statements is true according to the passage ?                  \\ \hline
    Answer            & \textbf{One night the writer forgot to lock the front door.}                \\ \hline
    BDG        & The writer couldn't find a hair cut in the barber shop.                        \\ \cline{2-2}
    \multirow{3}{*}{} & The writer couldn't find a hair cut in the shop.                         \\ \hline
    BDG ensemble       & The writer didn't want to open the front door.    \\ \cline{2-2}
    \multirow{3}{*}{} & The writer couldn't find the reason why he left the front door.                     \\ \hline
                      
    \end{tabular}
}
\caption{Yet another example for BDG multiple distractor generation}
\end{table*}

\end{document}